%% file: main.tex
\documentclass[10pt,twocolumn,letterpaper]{article}

\usepackage{cvpr}              

\input{preamble}

\definecolor{cvprblue}{rgb}{0.21,0.49,0.74}
\usepackage[pagebackref,breaklinks,colorlinks,citecolor=cvprblue]{hyperref}

\title{DiffAge3D: Diffusion-based 3D-aware Face Aging}

\author{
    Junaid Wahid$^{1,2}$, Fangneng Zhan$^{1}$, Pramod Rao$^{1}$, Christian Theobalt$^{1,2}$ \\
    \small $^1$MPI for Informatics
    \small $^2$Saarland University \\
    {\tt\small \{jjunaid,fzhan,prao,theobalt\}@mpi-inf.mpg.de}
}

\begin{document}


\twocolumn[{%
\renewcommand\twocolumn[1][]{#1}%
\maketitle
\includegraphics[width=\linewidth, height=5cm]{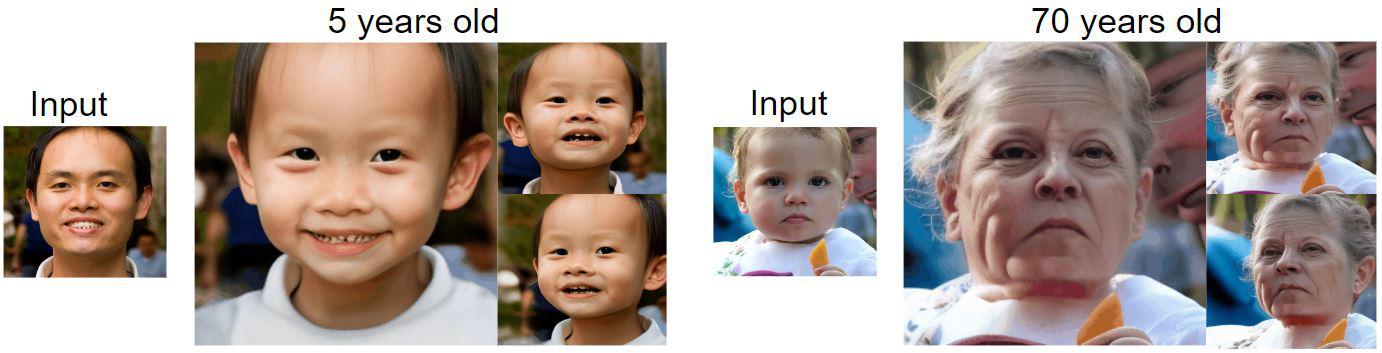}

\captionof{figure}{Given an input image (left) and target age(written on top), our method can synthesize consistent multiview aging results. These results can be rendered at arbitrary camera angles. \vspace{1em}}
\label{fig:teaser}
}]

\input{sec/0_abstract}    
\input{sec/1_intro}
\input{sec/02_relatedwork}
\input{sec/03_method}
\input{sec/04_experiments}

\input{sec/05_conclusion}
{
    \small
    \bibliographystyle{ieeenat_fullname}
    \bibliography{main}
}
\input{sec/X_suppl}

\end{document}

%% file: preamble.tex
\usepackage{booktabs} 
\usepackage[ruled]{algorithm2e} 
\usepackage{multirow}
\usepackage{adjustbox}
\usepackage{xspace}
\usepackage{graphicx}
\usepackage{float}
\usepackage{tikz}
\usepackage{comment}
\usepackage{csquotes}
\usepackage{wrapfig}
\usepackage{color}
\usepackage{lscape}
\usepackage{pifont} 
\usepackage{environ}
\usepackage{array}  
\usepackage[normalem]{ulem}
\usepackage{subcaption}

\newcommand{\OurMethod}{\textit{DiffAge3D}}

%% file: sec/0_abstract.tex
\begin{abstract}

Face aging is the process of converting an individual's appearance to a younger or older version of themselves. Existing face aging techniques have been limited to 2D settings, which often weaken their applications as there is a growing demand for 3D face modeling. Moreover, existing aging methods struggle to perform faithful aging, maintain identity, and retain the fine details of the input images. Given these limitations and the need for a 3D-aware aging method, we propose DiffAge3D, the first 3D-aware aging framework that not only performs faithful aging and identity preservation but also operates in a 3D setting. Our aging framework allows to model the aging and camera pose separately by only taking a single image with a target age. Our framework includes a robust 3D-aware aging dataset generation pipeline by utilizing a pre-trained 3D GAN and the rich text embedding capabilities within CLIP model. Notably, we do not employ any inversion bottleneck in dataset generation. Instead, we randomly generate training samples from the latent space of 3D GAN, allowing us to manipulate the rich latent space of GAN to generate ages even with large gaps. With the generated dataset, we train a viewpoint-aware diffusion-based aging model to control the camera pose and facial age. Through quantitative and qualitative evaluations, we demonstrate that DiffAge3D outperforms existing methods, particularly in multiview-consistent aging and fine details preservation.
\end{abstract}

%% file: sec/1_intro.tex
\section{Introduction} \label{sec:intro}

Deep learning techniques have made it possible to alter a person’s apparent age to any arbitrary age, offering exciting applications in fields like computer graphics and entertainment. However, current aging methods are limited as they are 2D-based and do not account for camera viewpoint, which is essential for understanding character appearance from all angles in industries such as entertainment and advertising.

Regarding these limitations, we identified that a major issue in current methodologies is their training data. 
Due to scarcity of large-scale paired aging dataset at different time lapse of same identity, several works \cite{olaf2020lifespan, S2gan2019, liu2019attribute, song2019dual, wanf2018faceage} have attempted to address this by dividing the limited dataset into distinct age groups to better understand aging patterns. 
These methods often utilize Generative Adversarial Network(GAN)~\cite{fellow2014gan} to model and learn the distribution of aging data.

Another line of work relies on utilizing pretrained image models like StyleGAN~\cite{Karras19} to learn the aging distribution without paired aging data.
These works \cite{abdal2020styleflow, harkonen2020ganspace, shen2020interpreting, wu2020stylespace} have explored latent manipulation approaches to identify the aging direction within the latent space of StyleGAN.  However, due to the entangled nature of StyleGAN’s latent space, aging often inadvertently changes other features, such as glasses or beards. To address this, SAM~\cite{alaluf2021only} uses two separate encoders for inversion and aging, aiming to disentangle the latent subspace of StyleGAN2~\cite{Karras2019stylegan2}. Despite this, SAM still experiences significant quality and information loss due to inversion. Building on SAM, the Production-Ready Face Re-Aging method~\cite{zoss2022production} uses SAM-generated data to train a consistent video aging model but it is limited to only ages 20 to 65. Most aging methods like SAM trained on the FFHQ~\cite{karras2019stylebased} dataset, which predominantly includes ages 25 to 50. This results in poor performance in other age group. Given the imbalanced structure and limited scale of existing datasets, acquiring large-scale paired datasets holds the potential to address these issues and enhance aging accuracy.


To address the scarcity of data, we propose a robust 3D aging dataset generation pipeline. Our pipeline is robust and primarily utilizes pretrained models to guide the learning of ideal aging transformations. 
Different from others, we do not use any inversion method, as we train our method on data sampled from the 3D GAN latent space. 
As aiming at 3D-aware aging which requires a 3D-aware aging dataset, we employ EG3D~\cite{chan2021}, a state-of-the-art 3D GAN, as the model for dataset generation.
Additionally, we incorporated CLIP into our pipeline for its rich understanding and alignment between text descriptions and images. We found that simple prompts like ``A photo of a 20-year-old person'' can yield faithful aging effect.
Our pipeline can generate multiview consistent aged versions of the same person, utilizing clip guidance and semantically rich EG3D's latent space to achieve smooth transitions between ages, including modeling very young and old ages.  With access to a large-scale dataset, we need to choose an appropriate method to model the aging process. Given the well-known limitations of GAN-based methods, particularly regarding inversion accuracy, we have opted for diffusion-based methods.

Recently, diffusion-based methods have outperformed GANs on a variety of tasks like image generation~\cite{dhariwal2021diffusion} and shown significant promise in the domain of face aging.
Specifically, FADING~\cite{chen2023face} fine-tuned the Stable diffusion model~\cite{rombach2022high} and then utilized Null text inversion~\cite{mokady2023nulltext} and the prompt-to-prompt method~\cite{hertz2022prompt} to achieve the desired aging effect at inference. However, its optimization step during inference often yields undesired results and artifacts. Besides, it can only model 2D data, having no idea of camera poses.
To model 3D-aware aging, we propose our camera viewpoint-aware diffusion model which is responsible for learning the aging effect by taking input images along with target age. Our model not only performs aging accurately but also preserves background and identity information, as illustrated in \cref{fig:teaser}. To control the camera pose faithfully, we fine-tune Zero123~\cite{liu2023zero1to3} on data generated from EG3D latent space, and then combined it with Controlnet~\cite{zhang2023adding} structure to manipulate the viewpoints. 
To facilitate smooth and consistent viewpoint transition, we incorporate motion modules which consist of a series of temporal attention layers. 
\\
Overall our contributions are threefold:

\begin{itemize}
     \item A robust 3D-aware aging dataset generation pipeline, which allows to generate an endless amount of consistent multiview aging data. 
     \item A novel 3D-aware diffusion model, to control the age and camera pose separately and faithfully.
     \item Extensive qualitative and quantitative experiments show that our method outperforms the state-of-the-art aging methods in terms of aging accuracy and 3D consistency.
\end{itemize}

%% file: sec/02_relatedwork.tex
\section{Related Work}

\textbf{Face Aging.}
Face aging has been a well-studied topic in the last couple of years. Under the context of 2D-aware aging,
Generative Adversarial Networks (GANs)~\cite{fellow2014gan} are widely employed for face aging and have demonstrated impressive results, as shown in \cite{Li2018global, Li2020hierarchical, olaf2020lifespan, S2gan2019, liu2019attribute, song2019dual,wanf2018faceage, abdal2020styleflow, harkonen2020ganspace, shen2020interpreting, wu2020stylespace}.

Specifically, SAM~\cite{alaluf2021only} employs an aging encoder along with an inversion encoder to achieve aging results in the latent space of StyleGAN2~\cite{Karras2019stylegan2}. Similarly, DLFS~\cite{He2021disentangled} relies on StyleGAN2, extracting shape, texture, and identity information and then aging through the StyleGAN2 generator. CUSP~\cite{cusp} separates the style and content of the input image, allowing it to achieve structure preservation along with aging. Unfortunately, GAN-based methods are notorious for losing either identity or details from input images. 
Attention-based methods~\cite{Zhu2020look, Chandaliya2023AWGAN} have also been explored for applying face aging to specific regions of the face. Recently, diffusion-based methods have demonstrated superior performance in preserving details and achieving effective aging. FADING~\cite{chen2023face} fine-tunes pre-trained diffusion and utilizes Null optimization~\cite{mokady2023nulltext} and prompt-to-prompt~\cite{hertz2022prompt} to achieve good aging results.  \\
\textbf{Novel View Synthesis.} Neural Radiance Fields(NeRF)~\cite{Mildenhall2020nerf} learn 3D scene using 2D images. \cite{Zhang2022fdnerf, Hong2022headnerf} achieved impressive results in 3D face novel views synthesis by employing NeRF in their frameworks. Utilizing neural radiance field, 3D-aware GAN~\cite{Deng2022gram, Gu2022stylenerf, Xue2022giraffehd} have achieved superior performance in terms of quality and speed. These works primarily use StyleGAN2~\cite{Karras2019stylegan2} and enable 3D understanding through different representations such as tri-plane in EG3D~\cite{chan2021}. Building upon EG3D~\cite{chan2021}, EG3D-GOAE~\cite{Yuan2023makeencoder} and Triplanenet~\cite{ Bhattarai2023triplanenet} proposed 3D inversion frameworks. Unfortunately, similar to 2D inversion, 3D GAN based inversion methods fail to preserve fine details and are unable to render challenging scenarios. Recently, diffusion models have gained a lot of attention in novel view synthesis. Most of these works leverage 2D diffusion models to learn the dynamics of the 3D world. Zero-1-to-3~\cite{liu2023zero1to3} has utilized a stable diffusion model to synthesize novel views using only a camera matrix. Similarly, \cite{SHI2023mvdream, LIU2023syncdreamer, Liu2023one2345plusplus, Zheng2023free3d} proposed novel view synthesis methods based on diffusion. Unlike 3D GANs, these methods often excel at reconstructing fine details but struggle with producing consistent views especially in portrait novel view synthesis.

Our 3D diffusion aging framework integrates the fine detail reconstruction strengths of diffusion methods with the consistent novel view synthesis capabilities of 3D GAN-based approaches to perform faithful multiview aging.

%% file: sec/03_method.tex
\section{Method}




\begin{figure*}
    \centering
    \includegraphics[width=18cm]{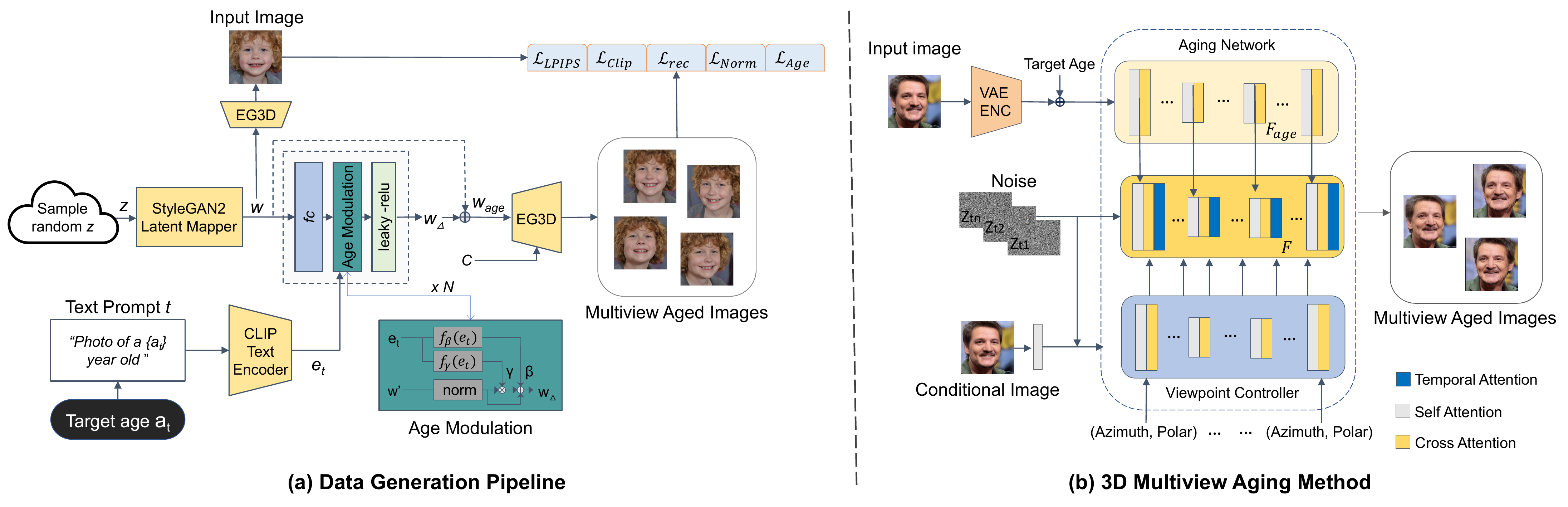}
    \caption{ 
    We present \OurMethod, a two-stage solution to solve the multiview aging problem. In the first stage, we train a data generation pipeline that trains a multiview aging generation pipeline without using any dataset. We sample latent vectors from StyleGAN2 and guide the aging process by utilizing a CLIP~\cite{Radford2021learning} model and pretrained age predictor. Lastly, we leverage a state-of-the-art 3D-aware Generative Model~\cite{chan2021} to obtain a faithful multiview aging result. In the second stage, we proposed a 3D diffusion-based aging framework. Our model can generate multiview aging results by only taking an input image and target age. We divided our model into 3 parts: Aging Network, Pose Controller, and Temporal Consistent Aging Module. Our whole method trains on a multiview synthetic aging dataset generated by our dataset generation pipeline.
    }
    \label{fig:pipeline_training}
\end{figure*}

Given a input image \(\boldsymbol{I} \in \mathbb{R}^{H \times W \times 3}\)\ and a target age \(\boldsymbol{T_{\text {age }}} \in \mathbb{R}^{1 \times 1}\)\ , our goal is to synthesize a multi-view consistent aging result. Our viewpoint control signal is defined by just two angles of azimuth \( \phi \) and polar \( \theta \):
\begin{equation}
\mathrm{I}_{\phi, \theta} = \mathrm{DiffAge3D}(\mathrm{I}, T_{\text{age}}, \phi, \theta).
\end{equation}

Where $\mathrm{I}_{\phi, \theta}$ is an aged version of a given input image \(\boldsymbol{I}\) at a newly synthesized viewpoint with azimuth \(\phi\) and polar angle \(\theta\).

Existing aging methods struggle with consistent aging and identity preservation. To address these issues, we propose a two-stage solution as illustrated in \cref{fig:pipeline_training}. First, we propose a robust data generation pipeline. We leverage the rich latent space of EG3D~\cite{chan2021} and further incorporate the powerful CLIP~\cite{Radford2021learning} text encoder to produce a diverse and high-quality aging dataset. Second, we propose a 3D diffusion-based aging framework for performing aging in a multiview setting. 

In the following sections, we first describe our data generation pipeline in Sec~\ref{subsec:data_pipeline}. Following that, we discuss our 3D diffusion-based multi-view aging method, which utilizes the dataset generated from our data generation
pipeline to model a 3D faithful aging, as detailed in Sec~\ref{subsec:3d_diff_model}.

\subsection{3D Aging Dataset Pipeline}
\label{subsec:data_pipeline}

We train our data generation pipeline without utilizing any available dataset. Specifically, we first randomly sample a noise vector \( z \in \mathbb{R}^{1 \times 512} \) and map it to the latent space of EG3D, \( w \in \mathbb{R}^{14 \times 512} \). Parallelly, we pass target age prompt \( t \) to the CLIP~\cite{Radford2021learning} text encoder to obtain text embedding \( e_t \). CLIP has rich text and image correspondence, allowing a simple age prompt like "Person of \{age\} years old" to guide the model in learning useful aging information.
We combine latent code and text embedding by introducing the Age Modulation Module, which takes \( w^{\prime}\) and \( e_t \) as inputs and results in an aging residual to be added to the original \( w \) vector. First, we pass the latent code \( w \) through \( \text{MLP}(w) \) to obtain \( w^{\prime}\). Then \( w^{\prime} \), along with the text embedding \( e_t \), passes through the following age modulation function:

\begin{equation}
w_{\triangle} = \left( f_{\gamma}(e_{t}) \right) \frac{w' - \mu_{\mathrm{w'}}}{\sigma_{\mathrm{w'}}} + f_{\beta}(e_{t}),
\end{equation}
where $w_{\triangle}$ is the residual direction for aging.  $\mu_{w^\prime}$ and $\sigma_{w^\prime}$ are the mean and standard deviation respectively, while $f_\gamma$ and $f_\beta$ are implemented using two fully connected layers. After obtaining the $w_{\triangle}$, we can add it to the originally sampled latent code $w$ to get the aged version of the sampled latent code:
\begin{equation}
    w_{age} =w +w_{\triangle}.
\end{equation}
Finally, we can render the $w_{age}$ using pretrained EG3D to get a consistent multiview aged output of sampled latent code.

\subsection{3D Diffusion-based Aging}
\label{subsec:3d_diff_model}
\subsubsection{\textbf{Aging Network}}

A faithful aging network should guide the aging process according to the target age while preserving the identity and other details from the input image. To achieve this, we introduced the Aging Network with the primary responsibility of ensuring faithful aging and maintaining identity. Previous studies relied on CLIP embeddings to encode identity information, but this often led to the loss of detailed information and struggle to encode information like age. Recently, MasaCtrl~\cite{Cao2023masactrl} introduced reference-based image conditioning, which preserves identity and low-level details while allowing transformations like viewpoint control. We followed a similar approach and made a few adaptations to develop a robust Aging network. 

Specifically, we propose an Aging Network that uses both the input image and the target age to generate an aged version of the input image. For the base network, we utilized Stable Diffusion~\cite{rombach2022high}, a latent diffusion model (LDM) denoted as \( F \), which we kept fixed throughout the training process. Our Aging Network, \( F_\text{age} \), is a trainable replica of the base network \( F \). We extract features from the Aging Network and use these features with the base network to produce the aged result during the denoising process. By keeping the base network's weights fixed and training only the Aging Network, we ensure that the model effectively learns to preserve identity. To further enhance the preservation of identity information, we avoid adding noise to the latent representation of the input image.
Additionally, we found that only concatenating the target age with the encoded latent of the input image as an extra channel facilitates the training of a robust Aging Network. 

The attention mechanism of base network \( F \) can be formulated as:
\begin{equation}
\text{Attention}(Q, K, V) = \text{Softmax}\left(\frac{Q K^{\intercal}}{\sqrt{d}}\right) V.
\end{equation}

The hidden attention states of Aging Network \( F_\text{age} \) can be denoted as \( \mathbf{y}_\text{age} \). We extract \( \mathbf{y}_\text{age} \) and pass them to the self-attention layers of the base network:
\[
\begin{aligned}
& \text{Attention}(Q, K, V, \mathbf{y}_\text{age}) = \text{Softmax}\left(\frac{Q K^{\prime T}}{\sqrt{d}}\right) V', \\
& Q = W^Q \mathbf{z}_t, \quad
K' = W^K \Vert \mathbf{z}_t, \mathbf{y}_\text{a}, \quad
V' = W^V \Vert \mathbf{z}_t, \mathbf{y}_\text{a},
\end{aligned}
\]
where $||$ denotes concatenation. The self-attention layers in LDM are responsible for maintaining the semantic layout of the image. By concatenating the features from the Aging Network, our model is able to perform aging by cross-querying content from both networks, allowing to learn important features necessary for generating an aged version of the input image while denoising.
\vspace{-4pt} 

\subsubsection{\textbf{Viewpoint Control}}

We have developed an Aging network that accurately transforms input images to target ages while preserving fine details. However, this network cannot currently control the viewpoint of the output images. Our goal is to introduce viewpoint control without modifying the Aging network’s weights.


To achieve this, we drew inspiration from two recent works. We integrated Zero-1-to-3\cite{liu2023zero1to3}, a view synthesis method using stable diffusion, which adjusts the viewpoint of the output image based on a camera matrix. However, Zero-1-to-3 struggles with consistency and generates poor results for novel face portraits. We improved it by fine-tuning with data sampled from EG3D, which enhanced consistency for novel portrait views. 


Currently, our Aging Network and fine-tuned Zero-1-to-3 are separate. We aim to combine them so that we can control the output's viewpoint without affecting the aging results. To do this, we plan to use a strategy similar to ControlNet~\cite{zhang2023adding}. ControlNet maintains the original network intact by creating a trainable copy and using a conditional image to guide the learning process.

We integrated the Aging network with our fine-tuned Zero-1-to-3 in a similar manner to the residual network training approach used in ControlNet. Specifically, we fixed the Aging network and initialized the trainable part of ControlNet with our fine-tuned Zero-1-to-3. However, we made two key modifications. Firstly, our fine-tuned Zero-1-to-3 takes the input image’s latent representation along with noise to preserve identity information, which is unnecessary in our framework because the identity information is provided by the Aging network. Secondly, although we use relative azimuth and polar angles to control the viewpoint, we still include a conditional image for ControlNet. Instead of adding an extra condition, we set the conditional image to be the input image, allowing the model to use it as a reference to capture fine details.

\subsubsection{\textbf{Temporal Consistent Aging Module}}
Our Aging Network ensures realistic aging, but our goal is to achieve consistent and realistic aging effects across all views. To achieve uniform aging and smooth transitions, we introduce the Temporal Consistent Aging Module. This module maintains temporal consistency, ensuring that the aging process remains coherent and continuous from any viewing angle.

Inspired by Animatediff\cite{Guo2023animatediff}, our method incorporates a trainable temporal module following the self-attention and cross-attention layers. Specifically, we inflated the original 2D-UNET to a 3D-UNET by introducing a new frame dimension in our input, represented as \( x\in \mathbb{R}^{B \times F \times C \times H \times W} \), where B denotes the batch size, F the frame, C the channel, and H and W the height and width, respectively.

To train the temporal layers while maintaining the integrity of other layers, we first reshape the input to \( x\in \mathbb{R}^{(B \times F) \times C \times H \times W} \), combining the batch size and frame dimensions into one dimension for processing by the image layers. Next, we reshape the input to \( x\in \mathbb{R}^{(B \times H \times W) \times C \times F} \) within the temporal layers. This second reshaping step allows the model to share information across different views, enabling cross-view attention and ensuring temporal consistency across all views.

%% file: sec/04_experiments.tex
\section{Results}
\label{sec:Results}

%


\begin{figure*}[t]
\centering
\includegraphics[width=0.86\textwidth,height=6.7cm]{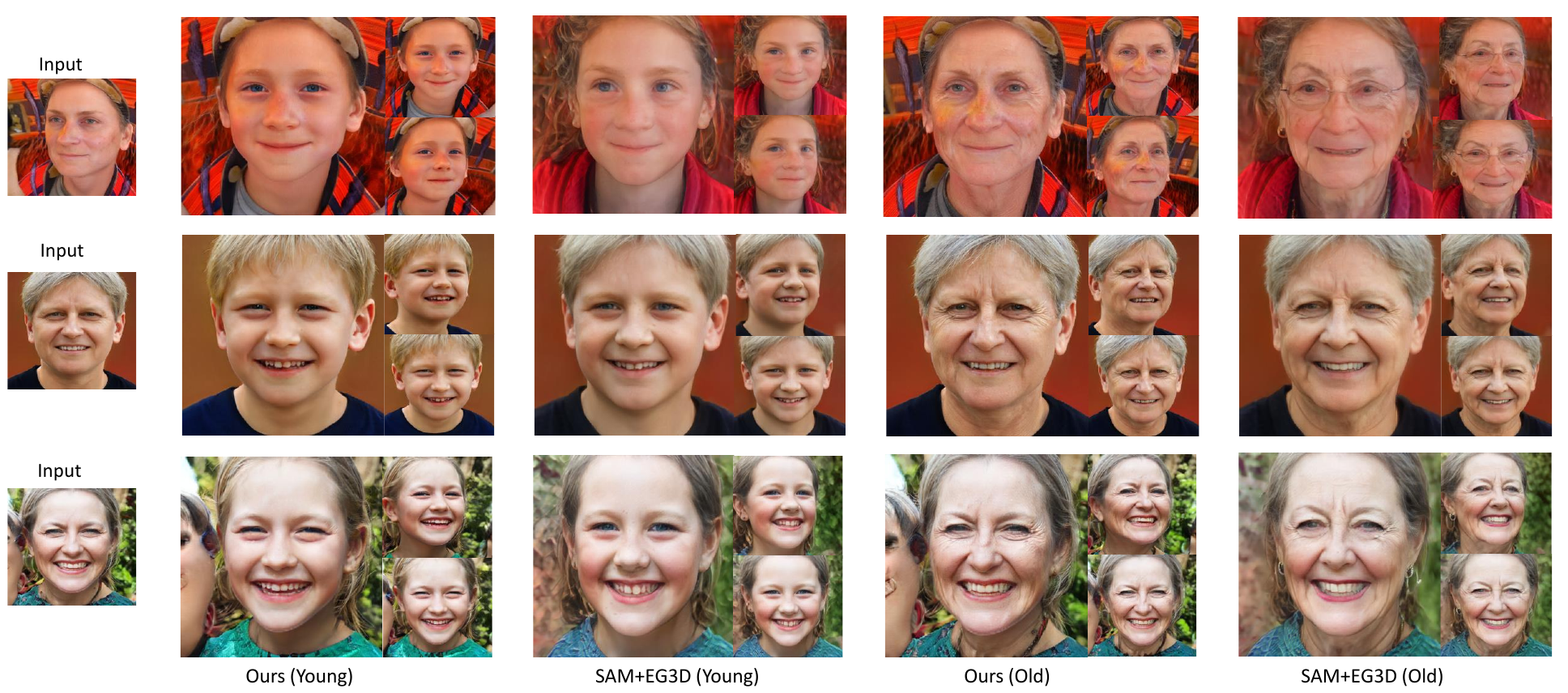}
\caption{Qualitative comparison of our data generation pipeline and SAM+EG3D. We compare both methods at two target ages: young (10 years old) and old (70 years old). Best viewed zoomed in}
\label{fig:data-gen-qualt}
\end{figure*}

\subsection{Training Details}
\label{subsec:training_details}

\subsubsection{\textbf{Data Generation Pipeline}}
We used a pretrained CLIP\cite{Radford2021learning} text encoder to obtain text embeddings and utilized pretrained EG3D\cite{chan2021} for rendering multiview images. Our losses of this stage are divided into two categories:

\textbf{Aging Losses}. We employ a pretrained age predictor, DEX\cite{Rothe2015dex}, to estimate the age of the generated aged output. We sample a random latent code $\mathbf{w}$ and the target age $\mathbf{a}_{\text{t}}$ to obtain $\mathbf{w}_{\text{age}}$. The reconstructed input image $\mathbf{x}$ is achieved from EG3D($\mathbf{w}$) and eventually, we generate aged image $\mathbf{x}_{\text{age}}$ by rendering EG3D($w_{\text{age}}$) from our pipeline.  
 We compute the $\mathcal{L}_{\text{2}}$ loss between the target age and the predicted age of the generated image.
\begin{equation}
\mathcal{L}_{\text{age}}\left(\mathbf{x}_{\text{age}}\right) = \left\|\alpha_t - \text{DEX}\left(\mathbf{x}_{\text{age}}\right)\right\|_2.
\end{equation}

We additionally employed directional clip loss $\mathcal{L}_{\text {clip }}$\cite{Patashnik2021styleclip} to guide the model to learn the aging effect from target age prompt $\mathbf{e}_{\text{t}}$ with the help of CLIP model:
\begin{equation}
\mathcal{L}_{\text{clip}} = D_{\text{CLIP}}\left(\mathbf{x}_{\text{age}}, \mathbf{x},e_{\text{t}}, e_{\text{i}}\right),
\end{equation}
where $\mathbf{e}_{\text{i}}$ is input age prompt and $D_{\text {CLIP }}$ is the CLIP-space cosine distance.

\textbf{Preservation Losses.}
Identity and fine details preservation are essential aspects of judging the quality of the generated aging dataset. To ensure pixel-wise similarity, we used $\mathcal{L}_{\text{2}}$ loss, and for perceptual similarity, we employed $\mathcal{L}_{\text{LPIPS}}$~\cite{Zhang2018unreasonable} loss. Besides, we employ norm loss $\mathcal{L}_{\text{norm}}$ to ensure the generation quality and utilized pretrained ArcFace\cite{Deng2019arcface}, denoted as $R$, to ensure identity preservation while aging:
\begin{equation}
\begin{aligned}
\mathcal{L}_{\text{ID}} = 1 - \langle R(\mathbf{x}), R(\mathbf{x}_{\text{age}}) \rangle
\end{aligned}
\end{equation}

%

The overall objective of our data generation pipeline is given as:
\begin{equation}
\begin{aligned}
\mathcal{L} = & \ \lambda_{\text{l2}} \mathcal{L}_2 + \lambda_{\text{lpips}} \mathcal{L}_{\text{LPIPS}} + \lambda_{\text{norm}} \mathcal{L}_{\text{norm}} \\
& + \lambda_{\text{id}} \mathcal{L}_{\text{ID}} + \lambda_{\text{age}} \mathcal{L}_{\text{age}} + \lambda_{\text{clip}} \mathcal{L}_{\text{clip}}
\end{aligned}
\end{equation}

where $\lambda_{\text{l2}},  \lambda_{\text{lpips}}, \lambda_{\text{norm}}, \lambda_{\text{id}}, \lambda_{\text{age}}, \lambda_{\text{clip}}$ are weights of the losses. We set $\lambda_{\text{l2}} =\text{0.1},  \lambda_{\text{lpips}} =\text{0.1}, \lambda_{\text{norm}} =\text{0.05}, \lambda_{\text{id}} =\text{0.2}, \lambda_{\text{age}} =\text{1}, \lambda_{\text{clip}} =\text{0.6}$. 

\subsubsection{\textbf{3D Diffusion Model}}
We employ a three-stage training strategy. In the first stage, we train the Aging Network independently of the viewpoint controller. The Aging Network is a trainable replica of LDM\cite{rombach2022high}, so we modify the input channel of it to accommodate an additional channel representing the target age.

In the second stage, we aim to train the viewpoint controller to control the pose of the aged image. First, we sampled a pair of data from EG3D, denoted as \(\{ (x, x_{\text{(azimuth, polar)}}, \text{azimuth}, \text{polar}) \}\),which include paired images and their corresponding azimuth \( \phi \) and polar \( \theta \) angles. We then fine-tune the Zero-1-to-3 model with the following objective:
\begin{equation}
\small
\min_{\theta} \mathbb{E}_{z \sim \mathcal{E}(x), \, t, \, \epsilon \sim \mathcal{N}(0,1)} \left\| \epsilon - \epsilon_{\theta} \left(z_t, t, c(x, \text{\( \phi \)}, \text{\( \theta \)}) \right) \right\|_2^2
\end{equation}
Here, Zero-1-to-3 comprises an encoder $\mathcal{E}$, a denoiser U-Net $\epsilon_\theta$ and a decoder $\mathcal{D}$.  The embedding of input images and angles is represented by $c(x, \phi, \theta)$. After fine-tuning, we freeze the aging network and initialize the trainable components of ControlNet with the fine-tuned Zero-1-to-3 model to fine-tune it further with the Aging network to control the pose of the aged image.

In the final stage, we freeze both the aging network and the viewpoint controller. We then train the temporally consistent aging module to ensure stable and consistent aging across all views. All training was conducted on one Nvidia A100 GPU with a learning rate of \( 10^{-5} \). During inference, we set 100 steps for DDIM denoising ~\cite{song2020denoising} and classifier guidance~\cite{ho2022classifier} as 3.


We trained all networks, except for fine-tuning the Zero-1-to-3, using data from our generation pipeline. We collected a total of 6,000 samples, with each example having 8 random views mapped to 3 target ages. This results in 24 images per input. No preprocessing was applied to the training data, and we set the image and text embeddings to Null in all networks.
\subsubsection{\textbf{Evaluation Metrics}} We compare our method with two state-of-the-art 2D aging methods: FADING\cite{chen2023face} and SAM\cite{alaluf2021only}. There are currently no existing multiview aging methods available for comparison. As a result, we extend 2D aging method's output to multiview using a pre-trained 3D inversion-based generative model, GOAE\cite{Yuan2023makeencoder}. In the following sections, we represented SAM\cite{alaluf2021only}+GOAE\cite{Yuan2023makeencoder} as SAM and FADING\cite{chen2023face}+GOAE\cite{Yuan2023makeencoder} as FADING for simplicity. We evaluate our method regarding aging accuracy, identity preservation, and viewpoint accuracy. Specifically, we use the pre-trained DEX model\cite{Rothe2015dex} to assess aging performance and employ SFace\cite{zhong2021sface} for identity preservation. Additionally, we utilize a pre-trained pose estimator\cite{Deng2019accurate} to estimate jaw and pitch for the generated images, computing the L2 loss based on these estimates to get viewpose results.

\begin{figure*}
    \centering
    \includegraphics[width=\textwidth, height=10cm]
    {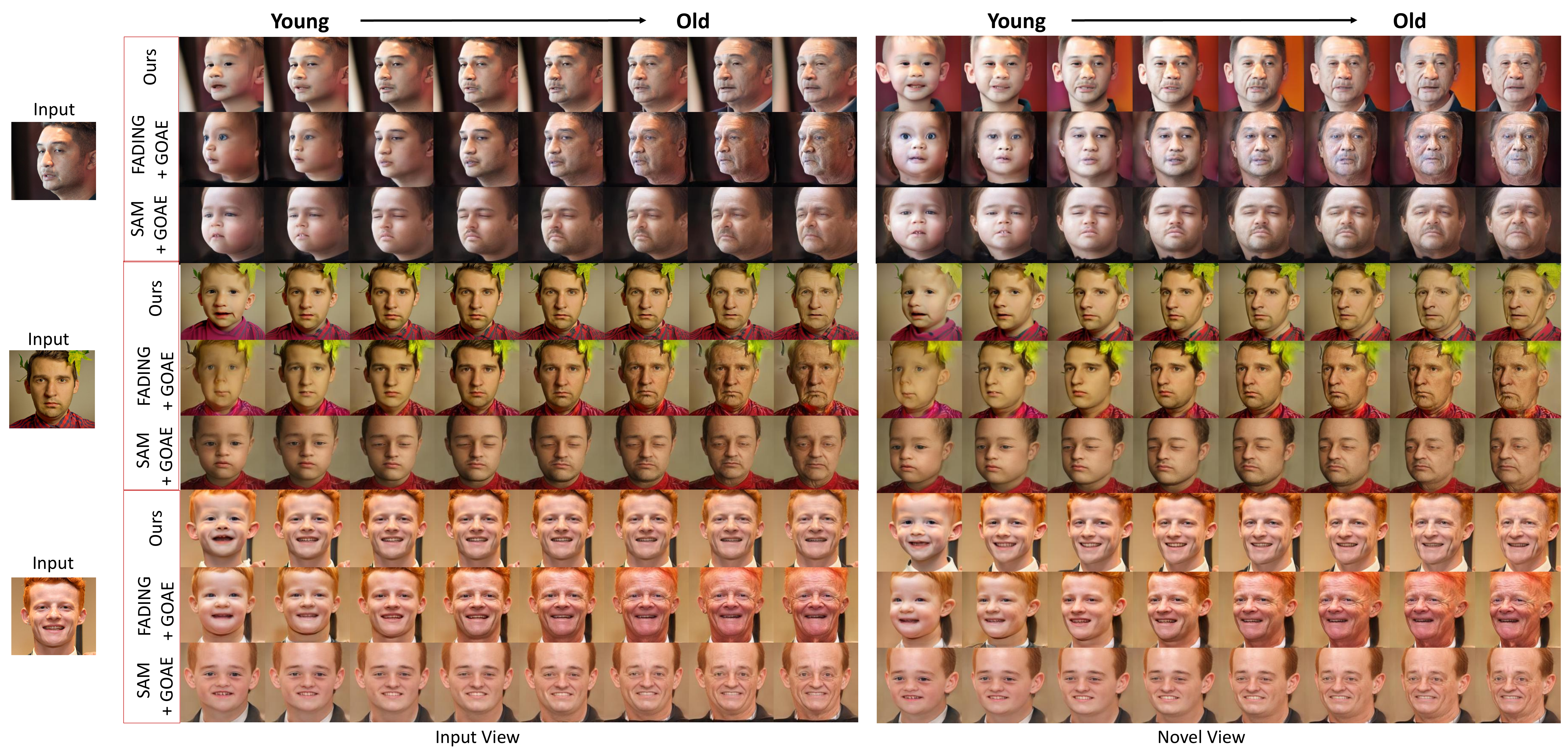}
    \caption{Qualitative comparison of aging results from 0 to 70 years at 10-year intervals. The left column shows the aging results on the input view, while the right column depicts aging results from a novel view. Best viewed zoomed in}
    \label{fig:aging_res}
\end{figure*}

\begin{table}
\renewcommand\tabcolsep{1.8pt}
\begin{tabular}{lcccc}
\hline
 &  Ours  &  FADING+GOAE &  SAM+GOAE   \\ \hline
 Age(MAE) $\downarrow$ & \textbf{7.46} & 7.62 & 8.80\\

 ID Preservation $\uparrow$ & \textbf{0.68} & 0.43 & 0.25 \\
 View direction $\downarrow$ & 0.0092 & 0.0021 & \textbf{0.0011}  \\ \hline
\end{tabular}
\caption{Quantitative analysis of method performances across metrics: Mean Absolute Error, ID Preservation, and view direction.} 
\label{tab:main_exp}
\end{table}

\begin{table}
  \centering
  \renewcommand\tabcolsep{1.8pt}
  \begin{tabular}{lcccc}
  \toprule
  & Ours & FADING+GOAE  & SAM+GOAE \\ 
  \midrule
  Age 0 to 30 & \textbf{25.670\(\pm\)7.19} & 45.79\(\pm\)13.88 & 18.19\(\pm\)6.16 \\
  Age 0 to 70 & 59.38\(\pm\)9.62 & \textbf{67.51\(\pm\)14.13} & 57.81\(\pm\)7.67 \\
  Age 70 to 30 & \textbf{29.31\(\pm\)5.92} & 28.80\(\pm\)6.16 & 29.23\(\pm\)5.93 \\
  Age 70 to 0 & \textbf{4.12\(\pm\)1.34} & 4.93\(\pm\)4.65 & 5.42\(\pm\)3.36 \\
  \bottomrule
  \end{tabular}
  \caption{Quantitative comparisons with baselines in challenging aging tasks. The table displays the mean predicted age along with the corresponding standard deviation.} 
  \label{tab:age_large_trans}
\end{table}

\begin{table}[h]
\begin{tabular}{lcc}
\hline
 &  Zero-1-to-3  &   EG3D-RGB    \\ \hline
 View Pose $\downarrow$ &  \textbf{0.0092} &  0.022\\

\end{tabular}
\caption{Ablation: Comparison of different viewpoint control options. The results are computed using pretrained pose estimator\cite{Deng2019accurate}} 
\label{tab:ablation_cond}
\end{table}

\begin{figure*}[h]
\centering
\includegraphics[width=0.96\linewidth,height=5.5cm]{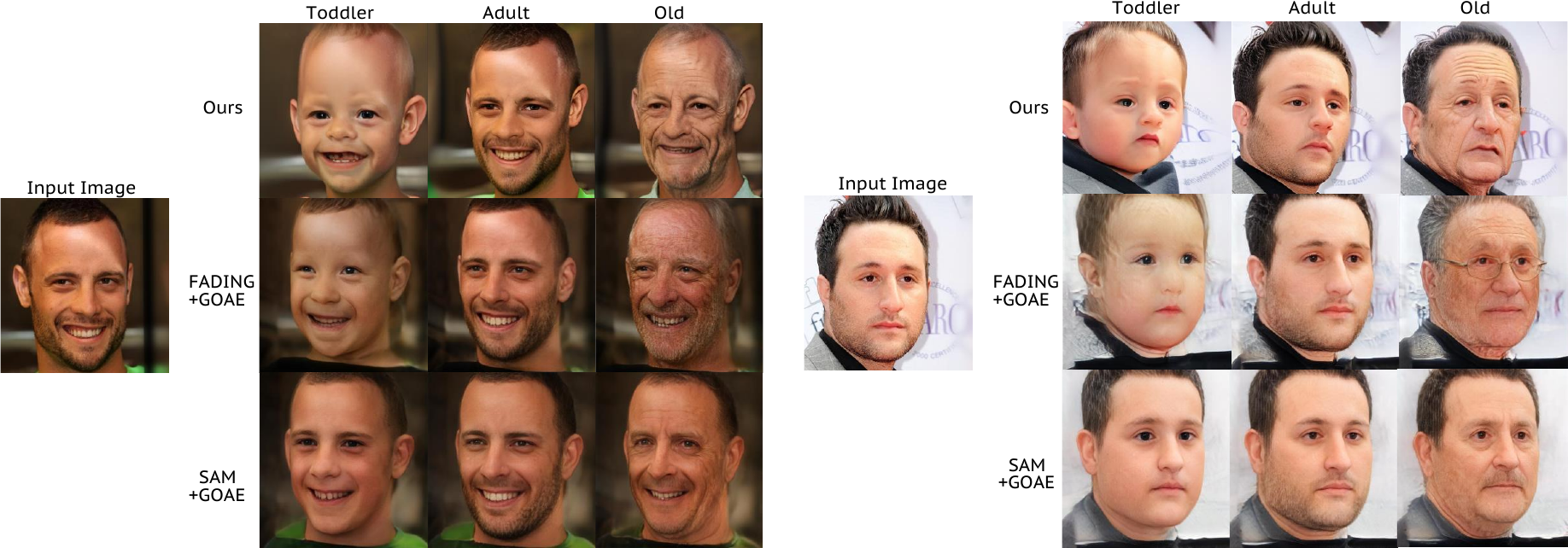}
\caption{Qualitative comparison of aging results generated from a novel view on the CelebA-HQ dataset across three age categories: Toddler (2 years old), Adult (25 years old), and Old (70 years old). Best viewed zoomed in}
\label{fig:celeb}
\end{figure*}

\begin{figure*}
\centering
\includegraphics[width=0.67\textwidth,height=4.4cm]{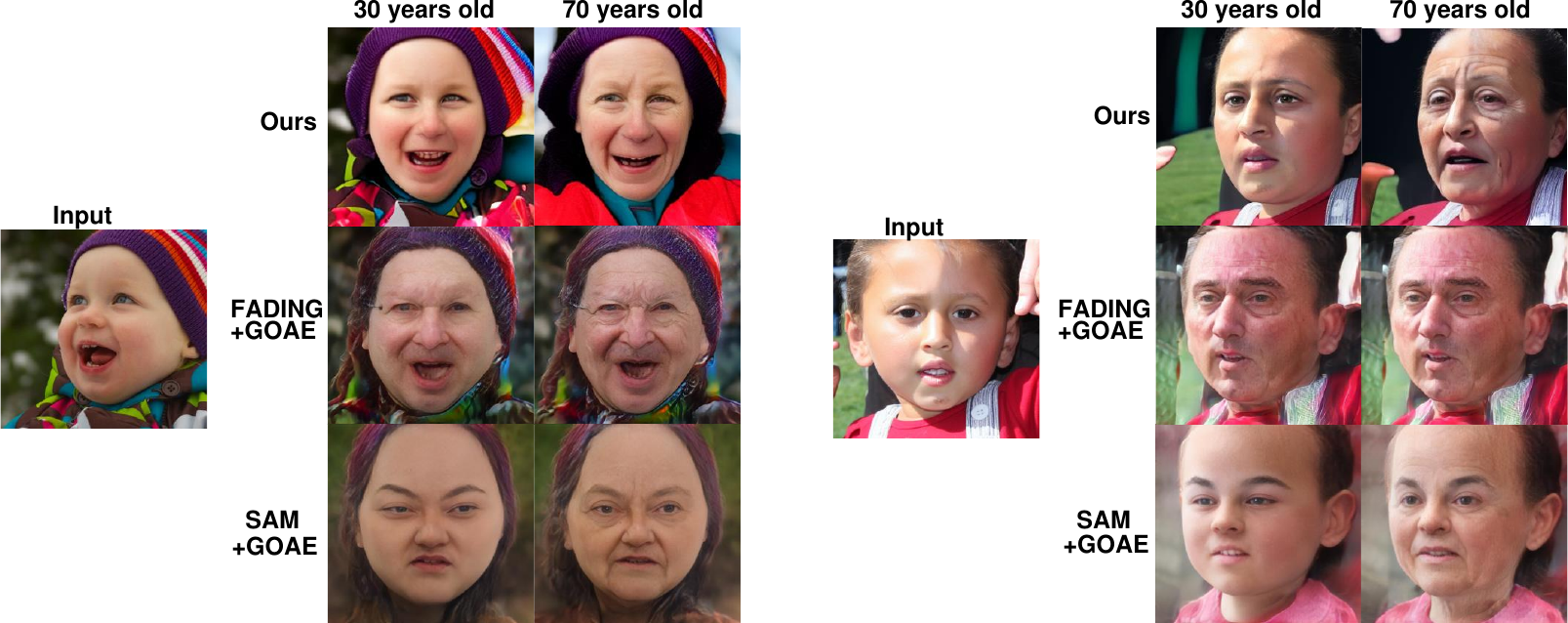}
\caption{ Qualitative comparison of challenging aging transformations: from toddler to
middle-aged(30 years) and elderly(70 years)}
\label{fig:age_large_trans}
\end{figure*}

\subsection{Data Generation Pipeline}\label{subsec:data_generator}

The training of diffusion models requires high-quality data. Our aging pipeline samples data from the EG3D latent space without an inversion bottleneck, ensuring no information loss. To evaluate whether our pipeline maintains identity and achieves accurate performance without inversion, we trained SAM\cite{alaluf2021only} with EG3D. We utilized all the same losses except for the CLIP loss. As shown in \cref{fig:data-gen-qualt}, our data generation pipeline not only performs aging accurately but also produces visually clearer and detailed results. Inversion-based methods often result in the loss of fine details, such as the cap in the first row of \cref{fig:data-gen-qualt}, and background artifacts visible in all examples of \cref{fig:data-gen-qualt}. Additionally, these methods may introduce unwanted objects like glasses, as observed in the old age example in the first row of  \cref{fig:data-gen-qualt}. Thus, our proposed method can effectively perform aging without comprising sample quality.

\subsection{Comparison with Face Aging Methods}\label{subsec:novel_view}
\vspace{-2pt} 
\textbf{Qualitative Comparison. }
To assess the effectiveness of our aging algorithm, we generated face aging results from ages 0 to 70 at 10-year intervals based on both the original input view and additional random novel view. The ~\cref{fig:aging_res} illustrates that our method consistently achieve accurate aging results across all age groups, including toddlers, while effectively preserving individual identity. DiffAge3D excels in shaping facial structures according to age, surpassing other methods. While FADING produces plausible facial shapes, it falls short compared to our approach. SAM struggles with maintaining identity and facial shape, often causing distortions. Our method retains background details better than FADING and SAM and preserves fine details more effectively due to the lack of reliance on pre-trained 3D generative models. We also compare our DiffAge3D against state-of-the-art methods on CelebA-HQ~\cite{lee2020maskgan} dataset in ~\cref{fig:celeb}. Our DiffAge3D outperforms all methods in both shape deformation and consistent aging (right subject in ~\cref{fig:celeb}). Additionally, our method effectively preserves light exposure and facial marks or bumps across all ages (left subject in ~\cref{fig:celeb}).

We further investigate the aging performance in challenging age transformation scenarios like toddler into a middle-aged or elderly person. The results in ~\cref{fig:age_large_trans} highlight our method's superiority over FADING and SAM. Our method performs aging smoothly and consistently. FADING fails at translating a toddler to a middle-aged person, and although SAM performs aging better, it does so at the cost of losing identity.



\textbf{Quantitative Comparison.}
We have observed better identity preservation and aging results over SAM and FADING in our qualitative comparisons. To validate these qualitative results, quantitative results for identity preservation, aging, and viewpoint are performed as shown in \cref{tab:main_exp}. Our model is trained on a synthetic dataset generated by our data generation pipeline. To better assess performance, we selected 200 random identities from the FFHQ dataset. Each test example was mapped to eight different age groups (0-70) at 10-year intervals. Since our method involves multiview 3D aging, we computed each metric on randomly selected eight different views.

The identity preservation metric is crucial for evaluating aging methods. As shown in \cref{tab:main_exp}, our approach outperforms all state-of-the-art methods, validating our qualitative results of consistent identity preservation. These results demonstrate that our data generation pipeline produces consistent multi-view aging results and our diffusion model effectively utilized this data to maintain identity consistency throughout the aging process.


Most state-of-the-art methods struggle with age translations across large age gaps, such as from toddlers to the elderly. To evaluate our model's performance on these challenging transformations, we collected 300 images of toddlers and 300 images of individuals aged 70+ from the FFHQ dataset. As shown in \cref{tab:age_large_trans}, our model performs better in most age group translations, such as transforming a toddler to age 30. However, for the transition from toddlers to the elderly, our results are comparable to other state-of-the-art methods. This may be due to the 3D model, GOAE\cite{Yuan2023makeencoder}, which might emphasize the visual appearance of wrinkles.

Additionally, we compared our performance on viewpoint control in \cref{tab:main_exp}. While other methods have a slight advantage here, we argue that their aging and identity preservation are not as robust as ours, as they do not significantly change the shape of the face structure.


\subsection{Ablations}\label{subsec:ablations}
\textbf{Viewpoint Controller.}
To control the pose of our output, we fine-tuned the Zero-1-to-3 model and integrated it into our viewpoint controller. To evaluate our choice of viewpoint controller, we conducted an ablation study, as detailed in \cref{tab:ablation_cond}.

As an alternative approach, we considered controlling the pose of the output using a random image.  Specifically, we aimed to sample a random image from EG3D and have our model replicate the pose from this image while preserving identity information from the Aging network. For this purpose, we employed ControlNet and used the sampled random image as the control condition.

The results, presented in \cref{tab:ablation_cond}, demonstrate that our method outperforms the alternative design in terms of viewpoint consistency. This improvement may be attributed to the fact that conditioning on an RGB image to control the viewpoint can sometimes interfere with aging features, potentially causing inaccuracies or artifacts in both viewpoint consistency and aging results.

%% file: sec/05_conclusion.tex
\section{Conclusion}
In this paper, we present \OurMethod, a two-stage approach to solving the task of 3D face aging. First, we propose a 3D-age-aware pipeline to address the scarcity of multiview aging datasets. Our robust 3D-age-aware dataset generation pipeline can generate high-quality multiview aging data using 3D GAN and CLIP's guidance. Later, we introduce a viewpoint-aware diffusion-based aging model to control the camera pose and age independently. Through extensive quantitative and qualitative evaluations, we demonstrate that our method offers significant advantages over existing state-of-the-art approaches, particularly in achieving 3D consistent aging and preserving identity. 



%% file: sec/X_suppl.tex
\clearpage
\setcounter{page}{1}
\maketitlesupplementary
\setcounter{section}{0}
\renewcommand{\thesection}{\Alph{section}}
\section{More Experiements results}

\subsection{Viewpoint Controller}
\begin{figure}

    \includegraphics[width=8.3cm]{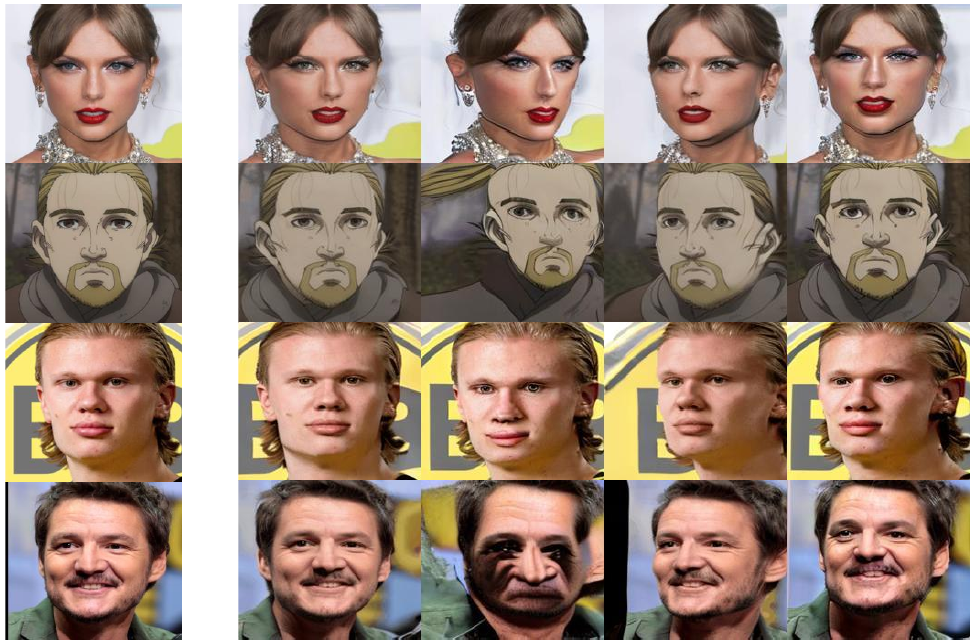}
    
    \vspace{-2pt} 

    {\fontsize{6.1}{8}\selectfont
    \hspace{9pt} 
    Input\hspace{38pt}Ours(View-1)\hspace{10pt}Zero123(View-1)\hspace{6pt}Ours(View-2)\hspace{6pt}Zero123(View-2)}

    \caption{Qualitative comparison of our fine-tuned Zero-1-to-3 and default Zero-1-to-3 models(represented as Zero123 in figure). We present novel view results for both methods on the faces of celebrities as well as out-of-distribution faces. Best viewed zoomed in}
    \label{fig:novel_view_zero123}
\end{figure}

We have fine-tuned the Zero-1-to-3 model to control the output pose and then integrated it using the ControlNet training strategy as our viewpoint controller. This integration allows precise control over the viewpoint of the aged output. As illustrated in \cref{fig:novel_view_zero123}, the fine-tuning has significantly improved the stability of Zero-1-to-3, ensuring that all samples closely follow the intended view directions. This is in contrast to the default Zero-1-to-3 model, which has difficulties with viewpoint consistency, often resulting in misalignment of facial features, such as the eyes. This fine-tuning enhances Zero-1-to-3's reliability as a zero-shot one-image-to-3D method and strengthens its role as a robust viewpoint controller in our multiview diffusion model.

\subsection{View-consistent novel view aging}
\begin{figure*}
    \centering
    \includegraphics[width=\textwidth, height=13cm]
    {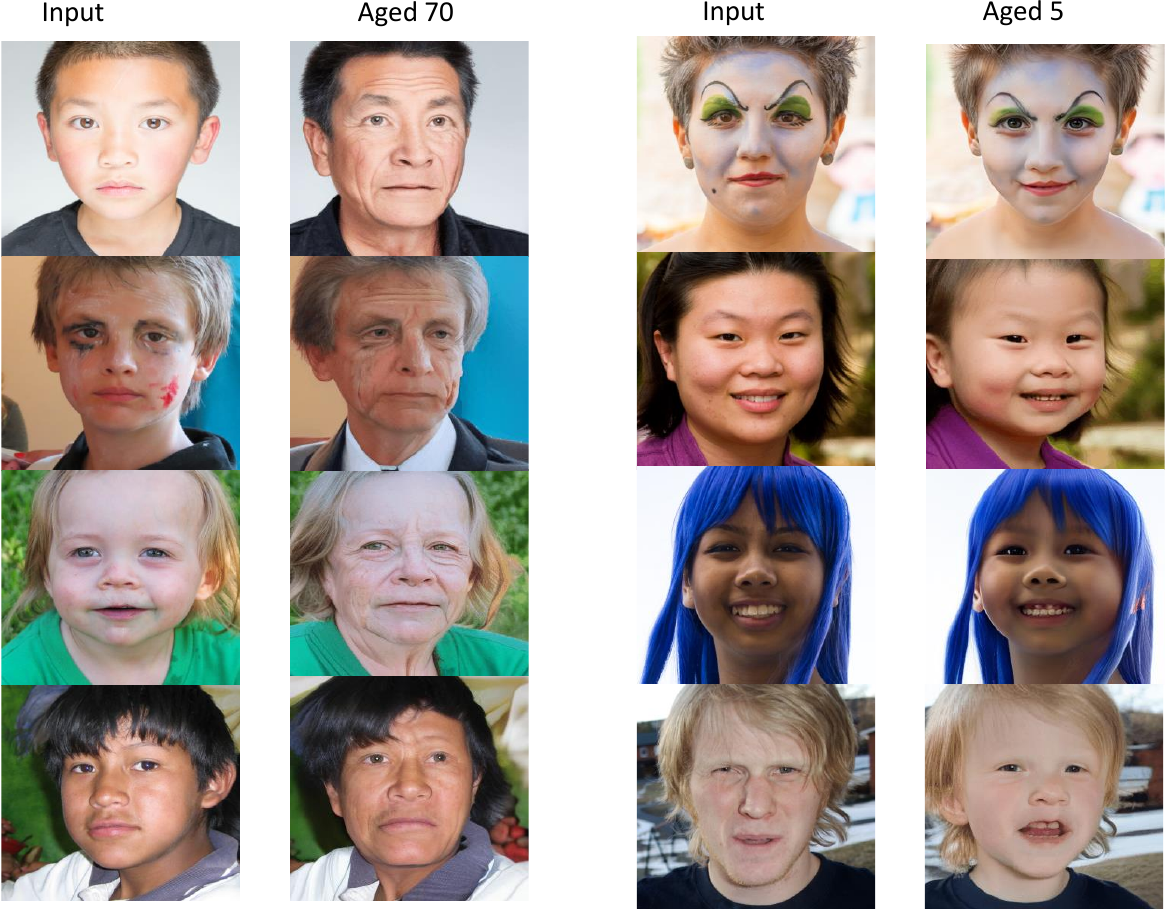}
    \caption{More Novel-view consistent aging results}
    \label{fig:aging_res_small}
\end{figure*}

\begin{figure*}
    \centering
    \includegraphics[width=\textwidth, height=18cm]
    {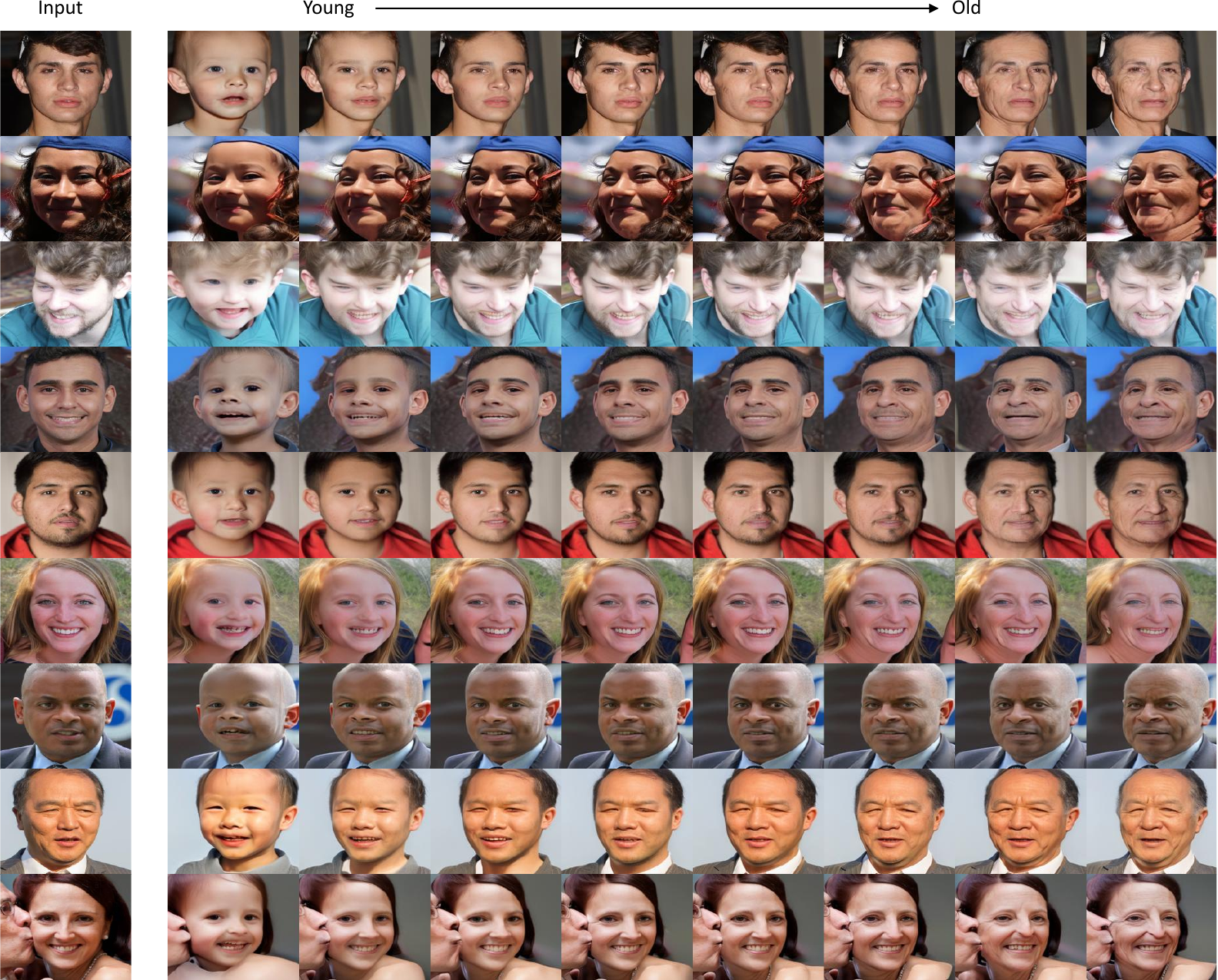}
    \caption{More Novel-view consistent aging results from ages 5 to 70 years. The results are generated from a random view.}
    \label{fig:aging_res1}
\end{figure*}

We present more challenging and impressive aging results in ~\cref{fig:aging_res_small}, and ~\cref{fig:aging_res1}. Our model effectively handles aging individuals with challenging poses, facial marks, and diverse races. It excels in aging across both directions, from toddler to elderly and elderly to toddler. For enhanced visualization of aging in a 3D setting, please refer to our supplementary video.

\section{Limitations and Future work}

\begin{figure*}[h] 
    \centering
    \begin{subfigure}[b]{0.46\textwidth}
        \includegraphics[width=8cm,height=4cm]{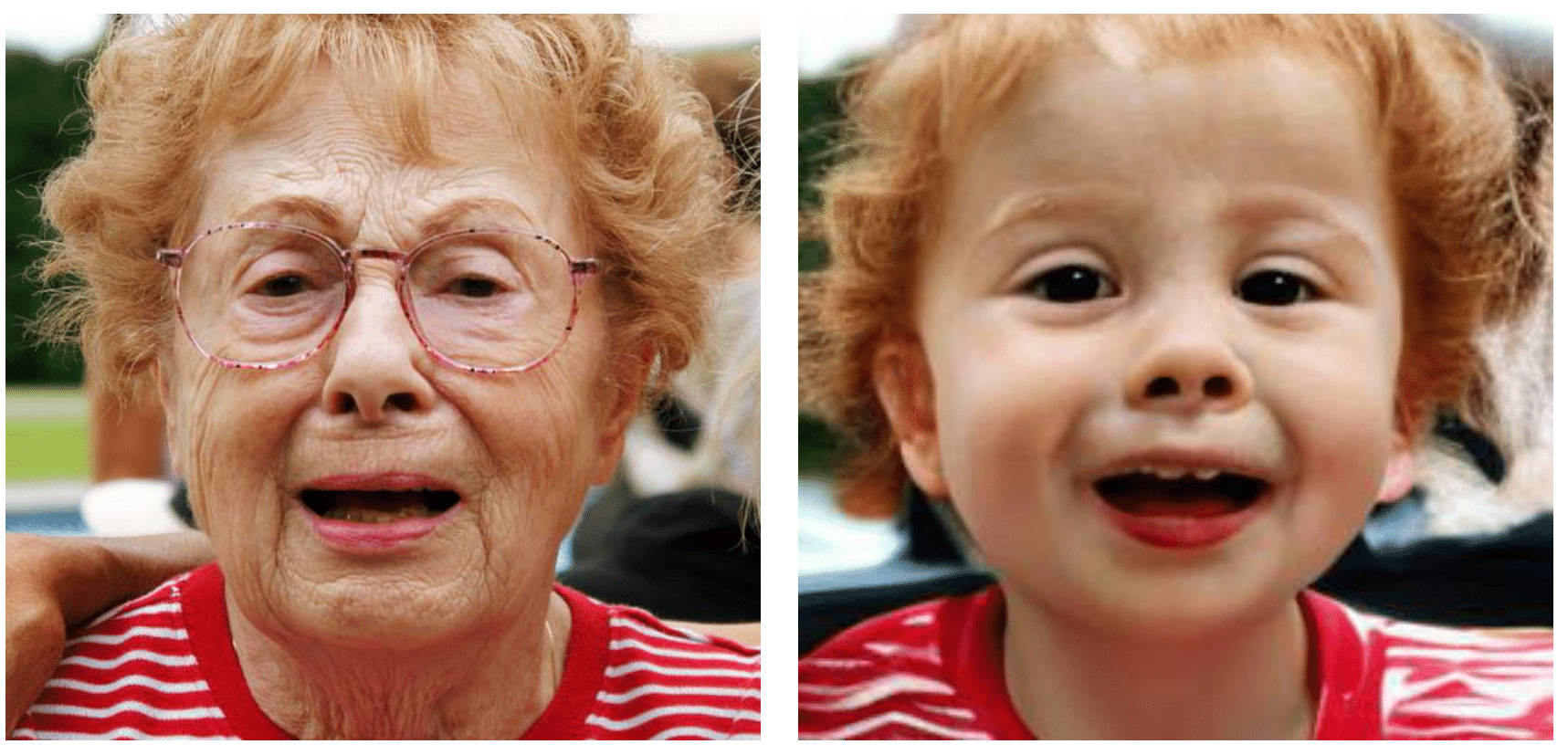}
        \caption{ Missing glasses in children}
        \label{fig:missing_glasses}
    \end{subfigure}
    \begin{subfigure}[b]{0.46\textwidth}
        \includegraphics[width=8cm,height=4cm]{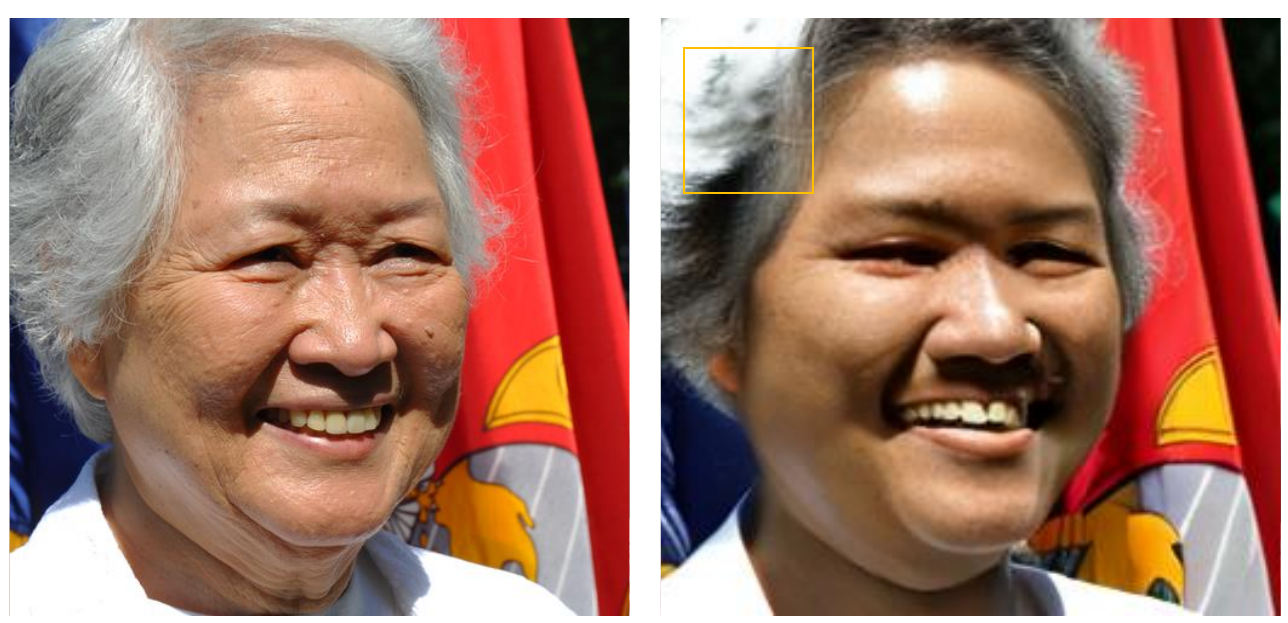}
        \caption{hair and background}
        \label{fig:hair_confusion}
    \end{subfigure}
    \caption{Limitation of DiffAge3D: Glasses sometimes disappear when old to toddler age transformation and  Misclassification of hair color as background in challenging scenarios}
    \label{fig:image_challenges}
\end{figure*}

While our method has achieved promising results, there are still some limitations that present opportunities for future research. One notable issue is that, when transforming images of older individuals into very young ones, glasses sometimes disappear. This problem, as shown in ~\cref{fig:missing_glasses}, arises from our data generation pipeline and is also observed in other aging methods like SAM and FADING. Moving forward, we aim to enhance our framework by incorporating text descriptions or more personalized approaches to provide detailed control over these subtle aspects.

Second, in particularly challenging scenarios, such as those shown in ~\cref{fig:hair_confusion}, where distinguishing between background and hair color becomes difficult, our model sometimes misclassifies these elements as background. Addressing these limitations is crucial and should be a focus of future research.

\section{Ethic Consideration}
We strongly discourage and condemn the misuse of generative AI for creating content that could harm individuals or spread misinformation. Our proposed framework could potentially be misused in facial recognition and identification systems. Therefore, we emphasize that this model is intended exclusively for legitimate and ethical research purposes, and we advocate for its use in generating positive and constructive content.
\vspace{10pt}


\vspace{13pt}
